\documentclass[letterpaper, 10 pt, conference]{ieeeconf}  

\IEEEoverridecommandlockouts                              

\makeatletter
\let\NAT@parse\undefined
\makeatother
\usepackage[pagebackref=false,breaklinks=true,colorlinks=true,bookmarks=false,linkcolor={red!50!black},urlcolor={magenta!80!black},citecolor={green!60!black}]{hyperref} 
\usepackage{url}
\usepackage{pifont}
\usepackage{booktabs} 

\usepackage{etoolbox}
\makeatletter
\patchcmd{\@makecaption}
  {\scshape}
  {}
  {}
  {}
\makeatletter
\patchcmd{\@makecaption}
  {\\}
  {.\ }
  {}
  {}
\makeatother

\usepackage{graphicx}
\usepackage{subfigure}
\usepackage{lipsum}
\usepackage{cite}
\usepackage{bm}
\usepackage{esvect}
\usepackage{amsmath}
\usepackage[ruled,vlined]{algorithm2e}
\usepackage{color}
\usepackage{amsmath,amssymb}
\usepackage{tabularx}
\usepackage{multirow}
\usepackage{arydshln}
\usepackage[dvipsnames]{xcolor}

\usepackage{overpic}
\usepackage{overpic} 
\usepackage{color}
\usepackage{mathtools} 
\usepackage{currfile}
\usepackage{multirow}

\definecolor{turquoise}{cmyk}{0.65,0,0.1,0.3}
\definecolor{purple}{rgb}{0.65,0,0.65}
\definecolor{dark_green}{rgb}{0, 0.5, 0}
\definecolor{orange}{rgb}{0.8, 0.6, 0.2}
\definecolor{red}{rgb}{0.8, 0.2, 0.2}
\definecolor{darkred}{rgb}{0.6, 0.1, 0.05}
\definecolor{blueish}{rgb}{0.0, 0.3, .6}
\definecolor{light_gray}{rgb}{0.7, 0.7, .7}
\definecolor{pink}{rgb}{1, 0, 1}
\definecolor{greyblue}{rgb}{0.25, 0.25, 1}

\renewcommand{\paragraph}[1]{\vspace{.5em}\noindent\textbf{#1}.}

\usepackage{lipsum}

\usepackage{blindtext}

\newcommand{\kostas}[1]{{\color{Bittersweet} {[\bf Kosta: #1]}}}
\newcommand{\andrew}[1]{{\color{BlueViolet} {[Andrew: #1]}}}
\newcommand{\vitto}[1]{{\color{OrangeRed} {[Vitto: #1]}}}
\newcommand{\JB}[1]{{\color{OliveGreen} {[Jon: #1]}}}
\newcommand{\tom}[1]{{\color{RoyalPurple} {[Tom: #1]}}}
\newcommand{\pratul}[1]{{\color{Emerald} {[Pratul: #1]}}}
\newcommand{\at}[1]{{\color{blueish}#1}}
\newcommand{\AT}[1]{{\color{blueish}{\bf [Andrea: #1]}}}
\newcommand{\At}[1]{\marginpar{\tiny{\textcolor{blueish}{#1}}}}
\newcommand{\al}[1]{\textbf{\color{orange}[AL: #1]}}
\renewcommand{\kostas}[1]{}
\renewcommand{\andrew}[1]{}
\renewcommand{\vitto}[1]{}
\renewcommand{\JB}[1]{}
\renewcommand{\tom}[1]{}
\renewcommand{\pratul}[1]{}
\renewcommand{\at}[1]{}
\renewcommand{\AT}[1]{}
\renewcommand{\At}[1]{}
\renewcommand{\al}[1]{}

\newcommand{\tabref}[1]{Tab.~\ref{#1}}
\newcommand{\equref}[1]{Eq.~\eqref{#1}}
\newcommand{\figref}[1]{Fig.~\ref{#1}}
\newcommand{\secref}[1]{Sec.~\ref{#1}}

\newcommand{\cmark}{\ding{51}}%

\makeatletter
\usepackage{xspace}
\DeclareRobustCommand\onedot{\futurelet\@let@token\@onedot}
\def\@onedot{\ifx\@let@token.\else.\null\fi\xspace}
 
\def\ie{i.e\onedot} 
 
\def\etc{etc\onedot}

\makeatother

\title{\LARGE \bf
MapNeRF: Incorporating Map Priors into Neural Radiance Fields \\ for Driving View Simulation
}

\author{Chenming Wu, Jiadai Sun, Zhelun Shen and Liangjun Zhang 
\thanks{
All authors are with Robotics and Autonomous Driving Lab (RAL), Baidu Research Institute, P.R. China. Z. Shen is the corresponding author.
{\tt\small \{wuchenming, sunjiadai, shenzhelun, liangjunzhang\}@baidu.com}}%
}

\begin{document}

\maketitle
\thispagestyle{empty}
\pagestyle{empty}

\begin{abstract}
Simulating camera sensors is a crucial task in autonomous driving. Although neural radiance fields are exceptional at synthesizing photorealistic views in driving simulations, they still fail to generate extrapolated views. This paper proposes to incorporate map priors into neural radiance fields to synthesize out-of-trajectory driving views with semantic road consistency. The key insight is that map information can be utilized as a prior to guiding the training of the radiance fields with uncertainty. Specifically, we utilize the coarse ground surface as uncertain information to supervise the density field and warp depth with uncertainty from unknown camera poses to ensure multi-view consistency. Experimental results demonstrate that our approach can produce semantic consistency in deviated views for vehicle camera simulation. The supplementary video can be viewed at \href{https://youtu.be/jEQWr-Rfh3A}{https://youtu.be/jEQWr-Rfh3A}.

\end{abstract}

\section{Introduction}

Autonomous driving (AD) 
has shown promising achievements and is considered an important technological breakthrough that could revolutionize the future of transportation. Currently, ensuring the safety of autonomous driving systems has become a topic of extensive development.
One traditional solution for safety tests is to exhaustively enumerate real scenarios for validation. Nevertheless, this process is not only labor-intensive and costly but also dangerous. Simulation has emerged as a robust, safe, and efficient alternative for training and evaluating AD software and algorithms~\cite{li2019aads, amini2020learning, amini2022vista}.

\begin{figure}
    \centering
    \includegraphics[width=\linewidth]{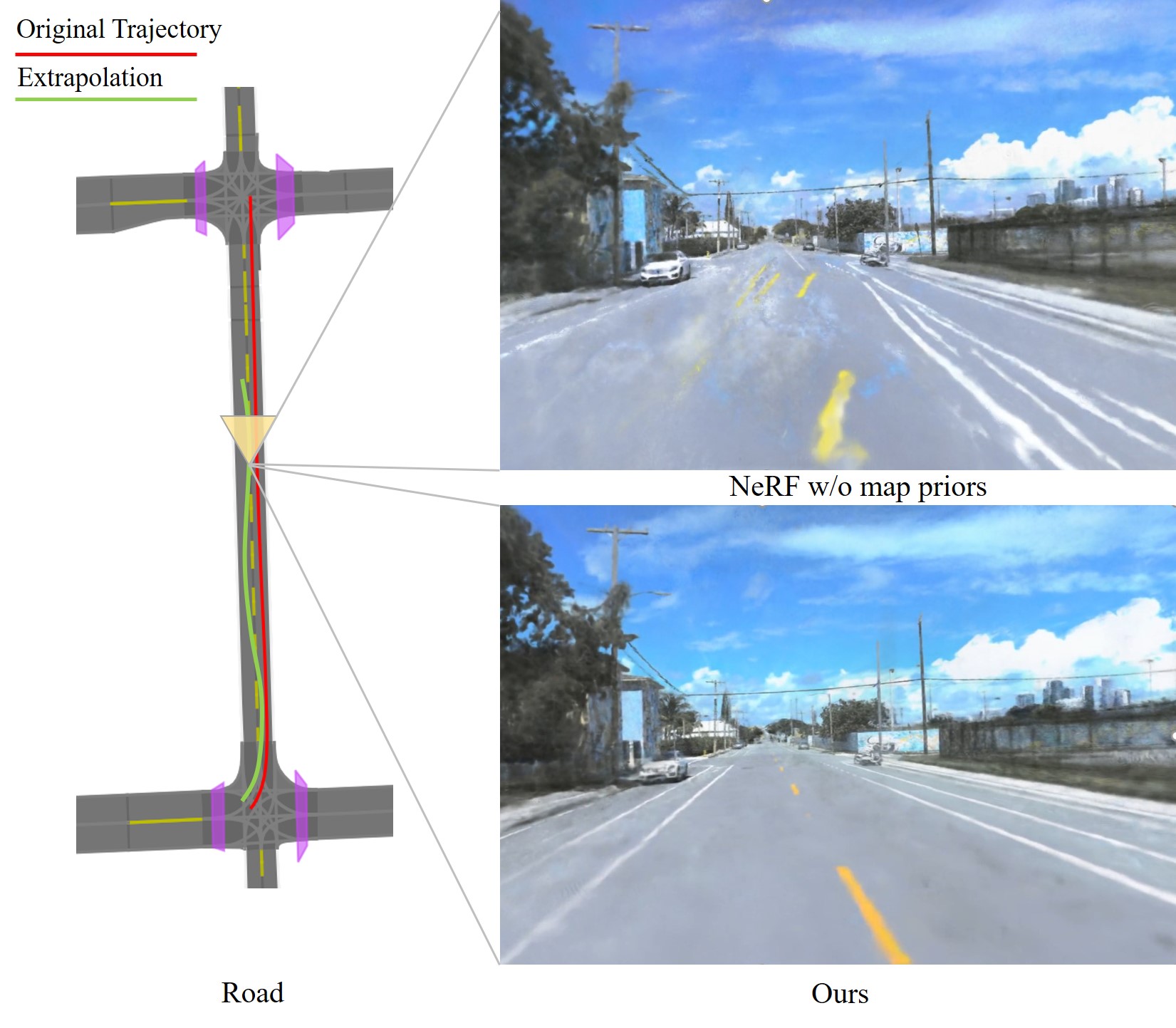}
    \caption{Simulating a driving view on the deviated trajectory (indicated in green). Our proposed method outperforms the existing NeRF method without map priors.}
    \label{figSupportComp}
    \vspace{-20pt}
\end{figure}

Recently, neural radiance field (NeRF)~\cite{mildenhall2020nerf} has gained significant attention in AD simulation~\cite{drivesim}. This approach leverages multi-view images to construct a 3D scene and enable novel view synthesis for both indoor and outdoor applications. When it comes to constructing NeRF models in AD simulation, there are two options available: 1) collecting a large amount of data to cover as many viewpoints as possible, and constructing a fine-grained scene offline; 2) directly using log data from road tests to quickly create an environment and dynamically simulate driving scenarios. The first choice can deliver high-quality simulation~\cite{tancik2022block} by transforming the problem of view extrapolation into view interpolation through the use of large amounts of data. However, it is time- and cost-intensive, which makes it challenging to generalize. As for the second choice, the collected images from log data are usually similar to each other along the running trajectory, which may result in unsatisfactory outcomes, particularly when the camera pose is placed out-of-trajectory (see \figref{figSupportComp} as an example), semantic consistency cannot be guaranteed when synthesizing images from deviated views. We observe this problem under this data condition in all neural radiance approaches, and to the best of our knowledge, none of the existing work has solved this issue.
In our opinion, semantic consistency is crucial for AD simulation, and synthesizing on deviated views is unavoidable for scalability.

AD simulation usually involves map data for planning and control, which can be obtained from a prebuilt High-Definition Map (HD Map) or an online mapping module. While the map data may not be pixel-perfect, it can provide semantic-level information that is useful for enhancing the semantic consistency of the trained neural radiance field.
In this paper, we propose incorporating map priors into neural radiance fields to enhance the semantic consistency and rendering quality of deviated driving view synthesis. Firstly, we employ ground information from maps to supervise the density field of NeRF, providing a more reliable road base for semantic entities. Next, we propose sampling rays to simulate unseen views. Unlike most NeRF augmentation methods~\cite{zhang2022ray, chen2022geoaug}, we utilize ground and lane information in sampling computations to guide the radiance field. More importantly, we model the above two supervision methods as weak supervision by using an uncertainty parameter and propose an uncertainty tempering scheme to increase the uncertainty. This ensures that map priors only guide the training process rather than enforce it towards their absolute values. As a result, our proposed method not only improves the rendering quality of interpolated novel view synthesis quantitatively but also enhances the semantic consistency of deviated novel view synthesis. 
Our contributions can be summarized as follows:

\begin{itemize}
    \item We propose a novel method to incorporate commonly used map priors in AD scenes into neural radiance fields to improve the out-of-trajectory driving view synthesis.
    \item We explicitly model the uncertainty in map priors as a parameter and propose an uncertainty tempering scheme to guide the training process of the neural radiance field.
    \item Experiments demonstrated that the proposed method can improve the semantic consistency of out-of-trajectory views and the rendering quality of novel view trajectory interpolation.
\end{itemize}

Our proposed method is easy to implement, can be easily plugged into existing NeRF algorithms, and has the capability of extending to other formats of priors.
\section{Related Work}

    

\subsection{Driving View Simulation}
The use of AD simulation has become increasingly popular in recent years, as it provides a valuable tool for verifying planning and control systems, synthesizing data for training and testing, and significantly reducing the time required for these tasks. There are currently two major types of simulators in use: model-based and data-driven. Model-based simulators, such as PyBullet~\cite{coumans2016pybullet}, MuJoCo~\cite{todorov2012mujoco}, and CARLA~\cite{dosovitskiy2017carla}, use computer graphics techniques to simulate vehicles and environments. However, creating these models and vehicle movements is often an expensive and time-consuming manual task, and the resulting images may not always be realistic enough, which can lead to degraded performance in deploying perception systems. Data-driven simulators, like AADS~\cite{li2019aads} and VISTA~\cite{amini2022vista, amini2020learning}, overcome these issues by using real-world datasets to create photorealistic simulations that are fully annotated and ready for training and testing of autonomous driving systems. Their driving view synthesis algorithms are largely based on conventional projection-based methods. A concurrent work UniSim~\cite{yang2023unisim} proposes an end-to-end simulation system using NeRF. Our method also builds upon the NeRF technique, which is capable of synthesizing photorealistic images, outperforming conventional view synthesis algorithms.

\begin{figure*}[!t]
    \centering 
    \includegraphics[width=\textwidth]{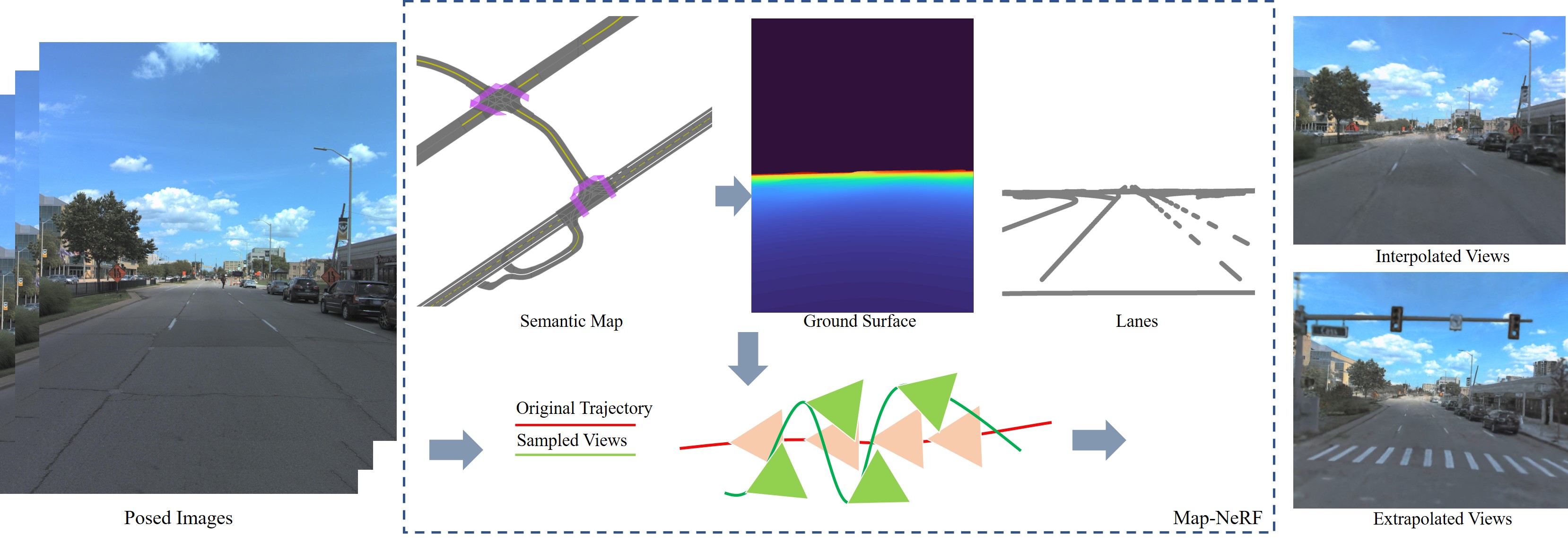}
    \caption{An overview of our proposed Map-NeRF method. Our approach involves incorporating map priors from semantic maps to guide the NeRF training process. This leads to better performance on deviated views, in terms of both semantic consistency and rendering quality.}
    \label{figSysPipeline}
\end{figure*}

\subsection{Neural View Synthesis (NVS)}
NVS is a long-standing task in the vision community, with significant potential for use in robotics. Recent surveys of methods can be found in~\cite{tewari2020state,tewari2021advances}.
Traditional novel view synthesis methods can be classified into three categories: image-based~\cite{gortler1996lumigraph,levoy1996light}, learning-based~\cite{niklaus20193d,rockwell2021pixelsynth}, and geometry-based~\cite{riegler2020free,riegler2021stable} approaches. 
Recently, neural implicit representations have shown promise for novel view synthesis.
NeRF~\cite{mildenhall2020nerf} is a seminal work that learns a continuous function to map spatial coordinates to density and color using MLPs to model a 5D radiance field. Subsequent works have extended NeRF to various scenarios, such as non-rigid and dynamic scenes~\cite{park:nerfies,park:hypernerf,Guo_2022_NDVG_ACCV}, larger unbounded scenes~\cite{zhang2020nerf++,tancik2022block,xiangli2022bungeenerf,rematas2022urban,martin2021nerf}, relighting~\cite{boss2021nerd,srinivasan2021nerv,zhang2021nerfactor}, and generalization ability~\cite{chen2021mvsnerf,trevithick2021grf,yu2021_pixelnerf_cvpr21,wang2021_ibrnet_cvpr21}. Additionally, some methods have been developed to optimize neural rendering more efficiently, such as~\cite{rebain2021_derf_cvpr21,Reiser2021_kiloNeRF_iccv21}, which subdivides the scene into multiple cells for efficient processing, and~\cite{yu2021plenoxels,sun2021direct,muller2022instant}, which exploit voxel-grid representations to speed up the optimization of radiance fields.

For novel view synthesis of street view, Block-NeRF~\cite{tancik2022block} decomposes a scene into blocks and trains NeRF individually along with appearance embeddings, learned to pose refinement, and controllable exposure. Urban-NeRF~\cite{rematas2022urban} and S-NeRF~\cite{anonymous2023snerf} incorporate LiDAR observations to supervise NeRF's depth to deal with unconstrained geometry in street views. 
In READ~\cite{Li23READ}, a point cloud representation is used for large-scale driving scenarios, which require pre-processing and may have view inconsistency problems. Kundu et al.~\cite{kundu2022panoptic} further extend NeRF to learn 3D semantic information in outdoor scenes.
In contrast to previous work, our paper is the first to use HD Map to assist in training NVS models.

\subsection{Limited Input View Synthesis with Geometric Constraint}
To improve NeRF's performance with fewer training views, several methods have been proposed. 
DS-NeRF \cite{deng2022depth} utilizes depth supervision to optimize a scene with a limited number of images. SynSin~\cite{wiles2020synsin} proposes a technique for synthesizing single-image views by warping the depth map and image features to generate new views of a scene from a single input image. 
GeoAug~\cite{chen2022geoaug} proposes a data augmentation method for NeRF that is based on geometry constraints with implicit depth supervision. Since ground truth images are not available for novel views, the rendered images of a novel pose are warped to the nearby training view based on the predicted depth maps and relative pose to match the RGB image supervision. GeCoNeRF\cite{kwak2023geconerf} warps the image of the seen viewpoint to the adjacent unseen viewpoint through depth and supervises the consistency at the image feature level. FWD~\cite{cao2022fwd} utilizes explicit depths and point cloud renders for fast rendering and fuses the warped features from multiple images to synthesize new images.
Unlike existing methods that use unsupervised or augmentation approaches, our technique leverages map priors to guide the training of the neural radiance field.
\section{Preliminary and Problem Definition}

NeRF uses a differentiable model of volume rendering to represent a scene as a volumetric field. 
It can be built upon multilayer perceptrons (MLPs), neural graphic primitive (NGP) or voxel grid, \etc, to encode the scene, which can be represented as a 5D function that takes a 3D location ${\rm x}=(x, y, z)$ and 2D viewing direction ${\rm d} = (\theta, \Phi)$ as inputs:
\begin{equation}
    \sigma, {\rm \bf c} = F({\rm d}, {\rm x}).
\end{equation} 

Given a set of images $\{{I}_i\}$ and corresponding camera poses $\{P_i\}$, NeRF casts each pixel from $I_i$ as a ray defined by the camera intrinsics, and sample particles to describe how much it blocks or emits lights along the ray. The color $\widehat{\mathcal{C}}({\rm{\bf r}})$ and depth $\widehat{\mathcal{D}}({\rm{\bf r}})$ of a ray $\rm \bf r$ can be approximated by integrating the sampled particles along the ray as follows,
\begin{equation}
    \widehat{\mathcal{C}}({\rm{\bf r}}) = \sum_{i=1}^N T_i (1-{\rm exp}(-\sigma_i\delta_i)) \rm {\bf c}_i ,
    \label{nerf:color}
\end{equation}
\begin{equation}
    \widehat{\mathcal{D}}({\rm{\bf r}}) = \sum_{i=1}^N T_i (1-{\rm exp}(-\sigma_i\delta_i)) \sum_{j=1}^i\delta_j,
    \label{nerf:depth}
\end{equation}
\noindent where $T_i\!=\!{\rm exp}(-\sum_{j=1}^{i-1}\sigma_i\delta_i)$ denotes the accumulated transmittance along the ray from the first sample to $i$-th sample, which is short for transmittance. $(1-{\rm exp}(-\sigma_i \delta_i))$ denotes the alpha value of the current sample contributed to the rendered color and depth, and $\sigma_i$ is the density of sample $i$, $c_i$ is the predicted color of sample $i$, and $\delta_i$ is the distance from sample $i$ to its next sample $i+1$. We denote the probability of ray termination as $h_i = T_i (1-{\rm exp}(-\sigma_i\delta_i))$. To supervise the training of $F$, an L2 photometric reconstruction loss is used:
\begin{equation}
    \mathcal{L}_{rgb}= \sum_i {\mathop{\mathbb{E}}_{{\rm \bf r}\in I_i} {|| \widehat{\mathcal{C}}({\rm{\bf r}}) - \mathcal{C}^{gt}_i({\rm{\bf r}}) ||_2^2}},
\end{equation}
\noindent where $\mathcal{C}^{gt}_i({\rm{\bf r}})$ is the ground truth color of $\rm \bf r$ from image $I_i$.
\vspace{4pt} \noindent \textbf{Problem Definition.} In our problem of using the neural radiance field $F$ for driving view synthesis, we are given $\{{I}_i\}$ and $\{P_i\}$, along with a semantic map $\mathcal{M}$. We define view interpolation as the use of a continuous function $\mathcal{F}_{in}$ to interpolate the orientations and translations of $\{P_i\}$, and synthesize novel views at any point on $\mathcal{F}_{in}$. Similarly, view extrapolation is defined as synthesizing a novel view at a point outside of $\mathcal{F}_{in}$, but near its nearest point on $\mathcal{F}_{in}$, typically to simulate a lane change in driving scenarios. In this setting, most existing NeRF methods exhibit decent performance on view interpolation but fail on view extrapolation.
To be more specific, we constrain $\mathcal{M}$ only to contain ground height and lane vectors, which are the fundamental entities that exist in most map formats. Our goal is to improve the quality of view extrapolation by incorporating $\mathcal{M}$ into the training of $F$ while ensuring that the performance of view interpolation is still maintained.

\section{Method}
In this section, we first introduce two supervision methods, \ie, ground density and multi-view consistency supervision. Then, we elaborate on how to incorporate map priors into them. After that, an uncertainty term with an uncertainty tempering strategy is explicitly modeled to weakly supervise the training and avoid introducing errors from the semantic-level map that might harm the neural radiance field.


\subsection{Ground Density Supervision}
\label{sec:ground_density}

\begin{figure}
\centering
\includegraphics[width=\linewidth]{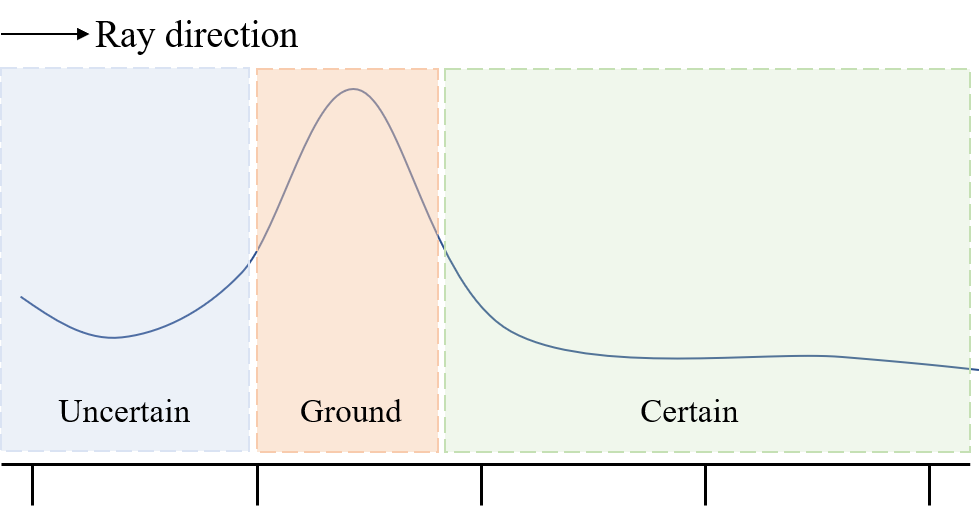}
\caption{The ray termination distribution of NeRF models. We partition the overall distribution into three regions: 1) uncertain region; 2) ground region; 3) certain region. The supervision should only be applied to the ground and 
 certain regions.}
\label{figRayDist}
\vspace{-10pt}
\end{figure}

The coarse geometry of HD map, in the format of the ground height field, motivates us to take it as an uncertain signal to guide the reconstruction process of the density field. We first analyze the formulation of ray termination (\ie \equref{nerf:depth}), which is a continuous probability distribution over the sampling region. As shown in~\figref{figRayDist}, we partition the ray termination distribution into three regions:
\begin{itemize}
    \item \textbf{Uncertain region}, where we are not confident if the ray should terminate in this region or not, due to the possible existence of obstacles or constructions on the road ground, which are not encoded in the map;
    \item \textbf{Ground region}, where the ground surface exists within an error tolerance (a typical average error is around $30cm$);
    \item  \textbf{Certain region}, or called as unconcern region, where the existence of any object beneath the ground is not related to our task of view synthesis.
\end{itemize}

Because a ray cast from images might terminate at the surface of obstacles or constructions on the road surface, we must not apply any supervision on the uncertain region. Therefore, we define the ideal distribution of ray termination as a multi-modal distribution instead of the unimodal distribution used in~\cite{deng2022depth}. Given the ground height map of $\mathcal{M}$, we first generate a dense triangle mesh $\widehat{\mathcal{M}}$ using Delaunay triangulation. Then, we use the camera intrinsic and extrinsic parameters to render $\widehat{\mathcal{M}}$ to the dense depth map $\mathcal{D}^{pgt}$ for each camera pose. Note that the term `\textit{pgt}' means pseudo ground truth. We use the KL divergence~\cite{deng2022depth} and the distant line-of-sight priors~\cite{rematas2022urban} to formulate our ground density supervision function:
\begin{equation}
\begin{aligned}
    \mathcal{L}_{gd} = &\mathop{\mathbb{E}}_{{\rm \bf r}\in I_i} \int_{\mathcal{D}^{pgt}_{ij}-\epsilon} \log h(t) \exp\left(-\frac{(t - \mathcal{D}^{pgt}_{ij})^2}{2\epsilon^2}\right) dt \\
    & + {\mathop{\mathbb{E}}_{{\rm \bf r}\in I_i} \int_{{\mathcal{D}^{pgt}_{ij}+\epsilon}} h(t)^2 dt},
    \label{eq:ground}
\end{aligned}
\end{equation}
where $\epsilon$ is the uncertainty that measures the upper bound of error between the map and the ground truth data. This integral can be easily approximated using a discrete set of samples.
To improve the smoothness of the learned depth images, we adopt the depth smooth regularization by sampling $2\times2$ patches rather than pixels to compute the gradients.

\subsection{Multi-view Consistency Supervision}
The ground density supervision enables the model to improve further the multi-view semantic consistency based on ray terminations. The key insight is ``\textit{a lane on the road should be a lane no matter where we look at it}''. We use a similar random ray casting method in~\cite{zhang2022ray} to generate rays from unseen views. However, the online pseudo labels generated by~\cite{zhang2022ray} are not suitable for our problem: the ray termination is not precise enough to be directly used for generating pseudo labels, though it has been explicitly supervised with uncertainty. In contrast, we define an uncertainty function to measure the confidence that a ray is terminated on the surface of the ground as:
\begin{equation}
    \Gamma({\rm \bf r_i}, \mathcal{D}^{pgt}_i) = \exp\left(\frac{-||\widehat{\mathcal{D}}({\rm \bf r_i}) - \mathcal{D}^{pgt}_i||_1}{2\epsilon}\right),
\end{equation}
\noindent where $\epsilon$ is the same uncertainty used in~\secref{sec:ground_density}. To stabilize the training process, we use ground depth $\mathcal{D}^{gt}_i$ provided by map priors to generate pseudo labels and use $\Gamma({\rm \bf r_i}, \mathcal{D}_i)$ to weight the gradients of rays. Concretely, for a ray cast ${\rm \bf r_i} = {\rm \bf o_i} + t \cdot {\rm \bf d_i}$ from $I$ and a sampled position ${\rm \bf o'_i} = {\rm \bf o_i} + \delta$, we use $\mathcal{D}^{pgt}_i$ to obtain a 3D point $p_i$ in global coordinate, and connect $p_i$ to $\rm \bf o'$ to obtain a sampled ray ${\rm \bf r'_i} = {\rm \bf o'_i} + t\cdot(({p_i-\rm \bf o'_i})/||{p_i-\rm \bf o'_i}||_2)$, where $\delta$ is a randomly sampled 3D vector with a norm of $0.1$ in our experiments. Optionally, we can use DPT~\cite{ranftl2021vision} to filter occluded regions on $\mathcal{D}^{pgt}_i$ out using least square fitting similar to~\cite{yu2022monosdf}. As a result, the ground truth of color $\mathcal{C}^{gt}_i$ can be used for supervising $\rm \bf r'_i$ with a weighting term $\Gamma({\rm \bf r_i}, \mathcal{D}^{pgt}_i)$. Additionally, if ${\rm \bf r_i}$ is passing through a pixel that is near rendered vector lanes, it can be enlarged according to the minimal distance. The loss of multi-view consistency can be written as:
\begin{equation}
    \mathcal{L}_{v} = \sum_i {\mathop{\mathbb{E}}_{{\rm \bf r'}\in I_i} { \Gamma({\rm \bf r_i}, \mathcal{D}^{pgt}_i)  || \widehat{\mathcal{C}}({\rm{\bf r'}}) - \mathcal{C}^{gt}_i({\rm{\bf r}})  ||_2^2}}.
    \label{eq:view}
\end{equation}

\subsection{Uncertainty Tempering}
The goal of our method is to use map priors to guide the training of neural radiance fields, formulating a weak supervision paradigm. The term $\epsilon$ used in the definitions of $\mathcal{L}_{gd}$ and $\mathcal{L}_{v}$ describes the uncertainty that we leave the radiance fields to explore, mainly using $\mathcal{L}_{rgb}$. In this section, we propose a strategy to enlarge uncertainty $\epsilon$ gradually, opposite to the simulated annealing of the learning rate frequently used in deep learning. This strategy provides more freedom for the radiance field as the training proceeds. We use the exponential tempering strategy to define the update equation for uncertainty tempering as follows.
\begin{equation}
    \epsilon' = \gamma \epsilon,
    \label{eq:ut}
\end{equation}
\noindent where $\epsilon'$ is the updated uncertainty, and $\gamma$ is the exponential growth rate (opposite to the decay rate), we set $\gamma =1.0005$ in our experiments.
The overall loss for training our semantic-consistency neural radiance field is:
\begin{equation}
    \mathcal{L} = \mathcal{L}_{rgb} + \lambda_d \mathcal{L}_{gd} + \lambda_v \mathcal{L}_{v}.
\end{equation}
where $\lambda$ is used to balance each loss, $\lambda_d = 0.2$ and $\lambda_v = 0.5$.

\section{Experiments}
We assessed the efficacy of our proposed approach in comparison to state-of-the-art NeRF methods. Our focus was on its capability to extrapolate beyond the original trajectory and interpolate novel view trajectories.


\begin{figure*}[!t]
    \centering 
    \includegraphics[width=\textwidth]{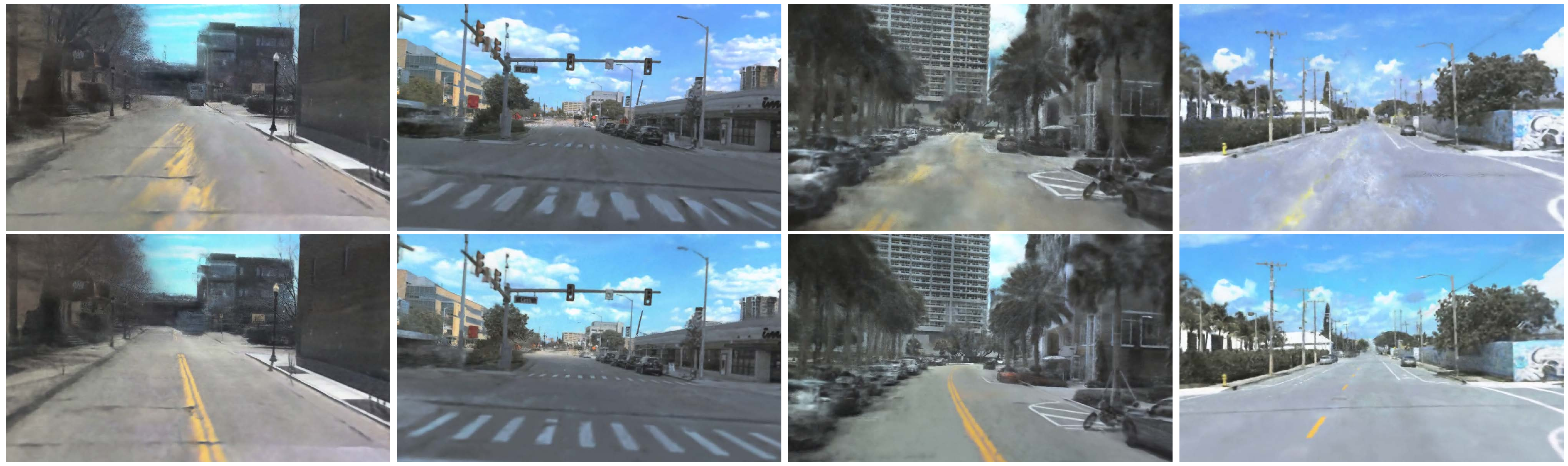}
    \caption{Synthesized images on deviated views. Our method has a significant improvement in road quality. 
    Top row: NeRF-base; Bottom row: Our Map-NeRF. 
    Best viewed in color and zoom in for details.
    For more visualizations, please refer to the supplementary videos.
    }
    \label{figExtrapolation}
\end{figure*}

\begin{figure*}[!t]
    \centering 
    \includegraphics[width=\textwidth]{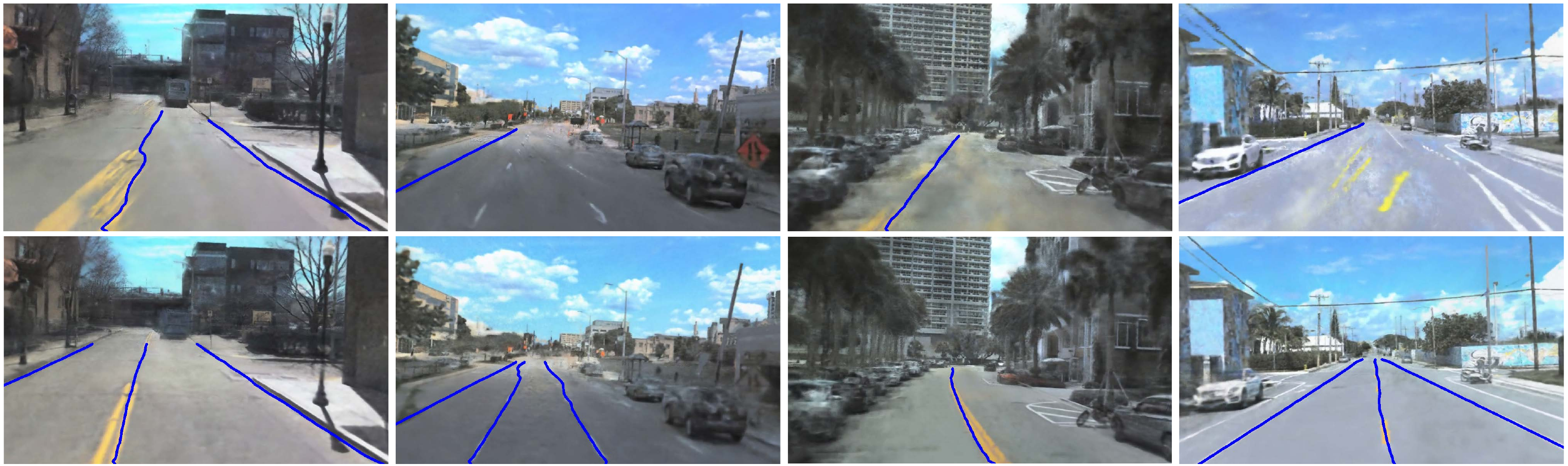}
    \caption{The results of lane detection (blue line) using the pre-trained CondLaneNet~\cite{liu2021condlanenet} model on 4 sequences from the dataset. 
    The new viewpoint images synthesized by our method can detect more accurate lane lines.
    Top row: NeRF-base; Bottom row: Our Map-NeRF.}
    \label{figLaneDet}
\end{figure*}

\subsection{Baselines}
\begin{table}[!t]
\vspace{6pt}
\caption{Evaluation of novel view trajectory interpolation.}
\label{tab:eval_nvs_metrics_interpolation}
\centering
\setlength{\tabcolsep}{7pt}
\renewcommand\arraystretch{1.2}

\begin{tabular}{cccc} 
\toprule
Methods & PNSR$\uparrow$  &  SSIM$\uparrow$ & LPIPS$\downarrow$  \\
\toprule
NeRF-base~\cite{tancik2023nerfstudio}  & 21.987 &	0.684 & 0.411 \\
\textbf{Ours}  & \textbf{22.353} &	\textbf{0.687} &   \textbf{0.410} \\

\bottomrule
\end{tabular}
\end{table}

\begin{table}[!t]
\caption{Evaluation of novel view trajectory extrapolation.}
\label{tab:eval_nvs_metrics_extrapolation}

\centering
\setlength{\tabcolsep}{7pt}
\renewcommand\arraystretch{1.2}

\begin{tabular}{cccc} 
\toprule
Methods & PNSR$\uparrow$  &  SSIM$\uparrow$ & LPIPS$\downarrow$  \\
\toprule
NeRF-base~\cite{tancik2023nerfstudio}  & 17.640 &	0.804 & 0.289 \\
\textbf{Ours}  & \textbf{19.702} &	\textbf{0.816} &   \textbf{0.273} \\

\bottomrule
\end{tabular}
\end{table}

Mip-NeRF 360~\cite{barron2022mip} achieves impressive performance on unbounded real-world scenes.
However, its training and testing speeds are relatively slower when compared to neural graphic primitive (NGP)-based methods~\cite{muller2022instant}. In our paper, we choose a recently proposed method -- Nerfacto~\cite{tancik2023nerfstudio} as our baseline, which combines Mip-NeRF 360's strengths and the advantages of Instant-NGP. For fairness, we opted to disable camera pose optimization and appearance embedding when presenting the evaluation metrics, as these could potentially impact computational results (optimized poses have no ground-truth images to evaluate). For simplicity, we refer to our utilized baseline method as ``NeRF-base". 
We run all experiments on a single NVIDIA Tesla V100 GPU. 
The initial learning rate is $2e^{-3}$ and decays to $1\times 10^{-4}$ for $30,000$ iterations in all experiments.

\subsection{Dataset and Metrics}

To evaluate the proposed method, we train MAP-NeRF on Argoverse2~\cite{wilson2023argoverse}, which has been specifically designed for autonomous driving. For our experiments, we utilized only the dataset's images, poses, and maps. To simulate forward driving scenarios, we only employed front-facing cameras (\ie left-rear, center, and right-rear). Our method and the baseline method were trained on six selected sequences from the Argoverse2 dataset, from which four of them are evaluated and visualized in the paper, and the others are shown in the supplemental video. Each sequence has $319$ frames, and the image size is $1550\times 2048$.

We evaluate every $9$ frames in the sequences, and the others will be used for the scene training.
To determine the pose of the images, we rely on the odometry poses provided by the dataset. To assess the quality of the synthesized images, we use a range of metrics that are commonly employed, including Peak Signal-to-Noise Ratio (PSNR), i.e., the $-log_{10}(\mathrm{MSE})$, Structural Similarity (SSIM)~\cite{wang2004image}, and Learned Perceptual Image Patch Similarity (LPIPS)~\cite{zhang2018unreasonable}. These metrics allow for a comprehensive evaluation of the image quality.

\subsection{Evaluation Results and Comparisons} 
\subsubsection{Novel View Trajectory Interpolation}
We compare our Map-NeRF with NeRF-base in terms of evaluation metrics on novel view trajectory interpolation.
As shown in \tabref{tab:eval_nvs_metrics_interpolation}, our method exhibits marginally better performance compared to the one of the NeRF-base.
For a qualitative comparison, please refer to the supplemental video.

\subsubsection{Out-of-Trajectory View Extrapolation}

\begin{figure}[!h]
    \includegraphics[width=0.95\linewidth]{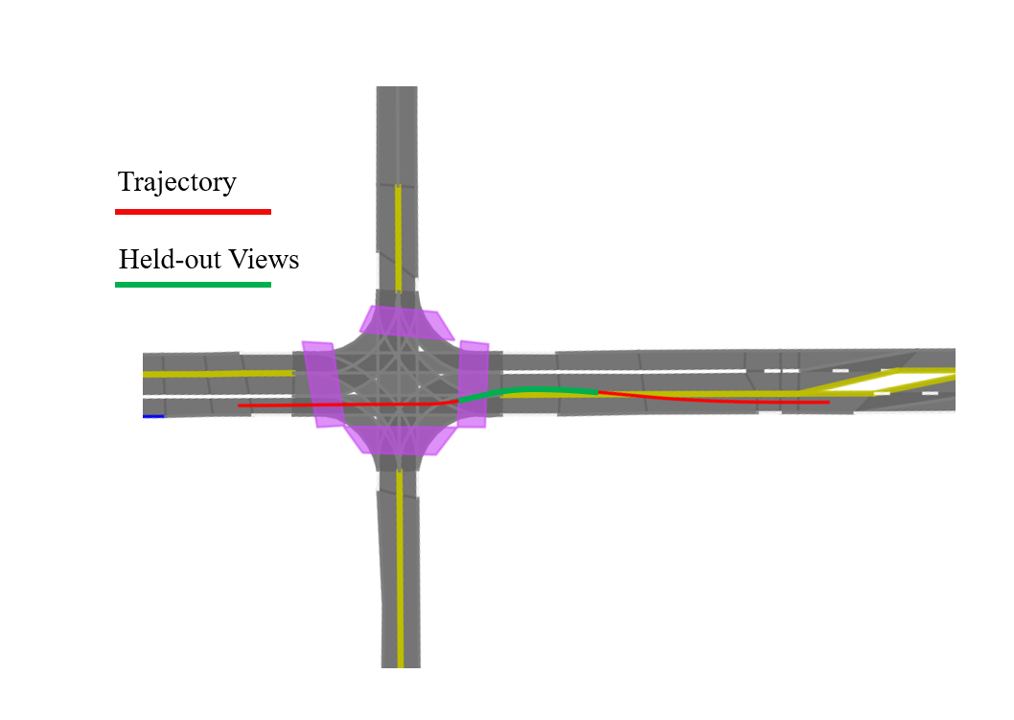}
    \caption{We conduct a quantitative evaluation of the synthesized images of deviated views, from a sequence containing lane changes in Argoverse2 dataset, Seq ID is $\rm 64b24fd1f6394f7ea535dbfe9fd737a1$.}
    \label{figAblation}
\end{figure}
This section presents both qualitative and quantitative results regarding deviated trajectory views. The qualitative results are shown in \figref{figExtrapolation}. We train our method against the baseline and render images on deviated views for qualitative evaluation.
Quantitative evaluation is challenging due to the lack of ground truth data. To address this challenge, we use a trajectory containing lane changes from the Argoverse2 dataset, as displayed in \figref{figAblation}. We hold out successive $50$ frames from the trajectory of lane changes for training and use them as ground truth images for quantitative evaluation. Experimental results refer to \tabref{tab:eval_nvs_metrics_extrapolation}. While this approach is not identical to view extrapolation, it approximates it and presents a challenging scenario for all NeRF methods. This can demonstrate the ability of our method for view extrapolation.

\subsection{Ablation Study}
We conduct several ablation experiments further to validate the effectiveness of the proposed three components, \ie, $\mathcal{L}_{gd}$, $\mathcal{L}_{v}$ and uncertainty tempering (UT in short). 
The ablation study is conducted on the same dataset we use for out-of-trajectory view extrapolation but with different combinations of the three components. 
As shown in \tabref{tab:loss_ablation}, our final setup (g) using all losses gains an improvement of $2.06$ PSNR compared to (a), which only uses $\mathcal{L}_{rgd}$ to supervise. As we add ground density supervision, multi-view consistency supervision, and uncertainty tempering, we see consistent improvements in our test scene. This indicates that our proposed methods are suitable ways to improve the deviated view synthesis.

\begin{table}
\vspace{6pt}
\caption{Ablation study on the loss function.}
\label{tab:loss_ablation}

\centering
\setlength{\tabcolsep}{7pt}
\renewcommand\arraystretch{1.2}

\begin{tabular}{cccccccc} 

\toprule
& {$\mathcal{L}_{rgb}$} & {$\mathcal{L}_{gd}$} & {$\mathcal{L}_{v}$} & {UT} & PNSR$\uparrow$  &  SSIM$\uparrow$ & LPIPS$\downarrow$  \\

\toprule
(a) & \cmark & &  &  & 17.640 &	0.804 & 0.289 \\
(b) & \cmark & \cmark &   &     & 18.346 &	0.806 &   0.287 \\
(c) & \cmark &   & \cmark &  & 19.408 &	0.810 &     0.280 \\
(d) & \cmark & \cmark & \cmark &  & 19.494&	0.811&       0.273 \\

\hdashline
(e) & \cmark & \cmark &   & \cmark& 18.407 &	0.808&     0.281\\
(f) & \cmark &   & \cmark & \cmark& 19.530 &	0.808&       0.279\\
(g) & \cmark & \cmark & \cmark & \cmark& \textbf{19.702}&	\textbf{0.816} & \textbf{0.273} \\
\bottomrule

\end{tabular}
\vspace{-6pt}
\end{table}


  



\subsection{Simulation Showcase}
This section provides an example of using our proposed method to simulate realistic visual sensors by rendering deviated views. We extract a few images from these views and apply a pre-trained lane segmentation algorithm, CondLaneNet trained on CULane dataset~\cite{liu2021condlanenet}, to obtain perception results. As shown in \figref{figLaneDet}, our proposed method produces more plausible and consistent results compared to the baselines. This demonstrates the potential of our method for use in building a data-driven AD simulator and its ability to produce semantically meaningful results.

\section{Conclusion and Future Work}

This paper proposed a new method to enhance out-of-trajectory driving view synthesis by incorporating commonly used map priors in autonomous driving scenes into neural radiance fields. To prevent the training process from collapsing due to the gap between map priors (semantic level) and neural radiance fields (pixel level), we explicitly model the uncertainty in map priors and propose an uncertainty tempering scheme to guide the process.
The results of our experiment show that our approach not only enhances the semantic consistency of views that deviate from the trajectory, but also improves the rendering quality of interpolating new views. Our method can be easily incorporated into most NeRF methods currently in use, and can be extended to other types of priors with the help of the uncertainty parameter.
Going forward, we aim to explore the incorporation of perception algorithms into self-driving cars, with the goal of enhancing real-time neural radiation fields. This could encompass various techniques such as semantic segmentation, lane detection, mapping, and dynamic obstacle in-painting.

\footnotesize
\bibliographystyle{IEEEtran} 
\bibliography{iros_abrv}

\end{document}